\documentclass[journal]{IEEEtran}
\usepackage [latin1]{inputenc}
\ifCLASSINFOpdf
\else
   \usepackage[dvips]{graphicx}
\fi
\usepackage{url}

\usepackage{graphicx}
\usepackage{url}
\usepackage{amsmath, amssymb}
\usepackage{mathrsfs}
\usepackage{amsopn}
\usepackage[ruled]{algorithm2e}
\usepackage{subfig, epsfig, epstopdf}
\usepackage{cite}
\usepackage[numbers,sort&compress]{natbib}

\begin{document}

\title{A Coupled Random Projection Approach to Large-Scale Canonical Polyadic Decomposition}

\author{Lu-Ming Wang, Ya-Nan Wang,  Xiao-Feng Gong, \IEEEmembership{Member, IEEE},  Qiu-Hua Lin, \IEEEmembership{Member, IEEE}, and Fei Xiang
\thanks{This work was supported in part by National Natural Science Foundation of China under grants 62071082, 61671106, 61871067, and in part by Fundamental Research Funds for Central Universities, China, under grant DUT20LAB120, DUT20ZD220. \textit{(Corresponding author: Xiao-Feng Gong.)}}
\thanks{Lu-Ming Wang, Ya-Nan Wang, Xiao-Feng Gong,  and Qiu-Hua Lin were with the School of Information and Communication Engineering, Dalian University of Technology, Dalian 116024, China (e-mail:12414029152@qq.com; wynhhh@mail.dlut.edu.cn; xfgong@dlut.edu.cn;  qhlin@dlut.edu.cn).}
\thanks{Fei Xiang, was with Xiaomi Inc. Artificial Intelligence Department, AI Lab, Beijing 100085, Chian (e-mail: xiangfei@xiaomi.com).}}

\maketitle

\begin{abstract}
We propose a novel algorithm for the computation of canonical polyadic decomposition (CPD) of large-scale tensors. The proposed algorithm generalizes the random projection (RAP) technique, which is often used to compute large-scale decompositions, from one single projection to multiple but coupled random projections (CoRAP).  The proposed CoRAP technique yields a set of tensors that together admits a coupled CPD (C-CPD) and a C-CPD algorithm is then used to jointly decompose these tensors. The results of C-CPD are finally fused to obtain factor matrices of the original large-scale data tensor. As more data samples are jointly exploited via C-CPD, the proposed CoRAP based CPD is more accurate than RAP based CPD. Experiments are provided to illustrate the performance of the proposed approach. 

\end{abstract}


\IEEEpeerreviewmaketitle

\section{Introduction}

\IEEEPARstart{W}{ith} the size and dimension of datasets growing much faster than ever, decomposition of large-scale datasets has become an important issue in signal processing and machine learning. As such, large-scale tensor decomposition has attracted much attention recently, and a number of algorithms\cite{vervliet2015randomized,phan2011parafac,sidiropoulos2014parallelA,erichson2020randomized,yuan2019randomized,che2019randomized,minster2020randomized} have been proposed so far. In the above-mentioned algorithms, various randomization methods play an important role, and on top of that, random projection (RAP)\cite{erichson2020randomized,sidiropoulos2014parallelA} has been widely adopted in the decomposition of large-scale tensors into various models, including canonical polyadic decomposition (CPD)\cite{sidiropoulos2014parallelA,erichson2020randomized}, Tucker\cite{minster2020randomized}, tensor train (TT)\cite{che2019randomized} and tensor ring (TR)\cite{yuan2019randomized}.

Besides the above works on large-scale tensor decompositions, coupled decompositions of multiple datasets, including coupled matrix-tensor decomposition\cite{acar2015data}, coupled and double coupled CPD\cite{sorensen2015coupledA,sorensen2015coupledB,gong2018double,gong2019double,gong2013joint,farias2016exploring} and coupled block term decomposition\cite{gong2016coupledTerm,yang2019using}, have also received much attention in the recent decade. These coupled decomposition techniques are important tools in multi-set data fusion and are shown to have better performance than their uncoupled counterpart with regards to both identifiability and accuracy.

In this paper, we combine the idea of RAP and that of coupled tensor decomposition to develop a novel algorithm for the CPD of a large-scale tensor. We will propose a new coupled RAP (CoRAP) approach that can generate multiple coupled sets of projection matrices, and applying these sets of projection matrices to the data tensor yields multiple compressed tensors. Due to the CPD structure of the original tensor and the coupling among different sets of projection matrices, the set of compressed tensors together admits a coupled CPD (C-CPD). Therefore, the compressed tensors can be jointly decomposed  via a C-CPD algorithm and the results can then be fused to obtain the factor matrices of the original large-scale data tensor. Noting that the proposed CoRAP based CPD algorithm exploits more structure, i.e., multiple compressed tensors as well as their coupling, than its RAP based counterpart, it is expected to have better performance with regards to accuracy. For convenience, we limit ourselves to third-order real-valued tensors in this paper, although the presented results can be analogously extended to complex-valued tensors with order higher than three.

\textit{Notations}: vectors, matrices and tensors are denoted by lowercase boldface, uppercase boldface and uppercase calligraphic letters, respectively. The $ r $th column vector and the $ (i, j) $th entry of $ \mathbf{A} $ are denoted by $ \mathbf{a}_r $ and $ a_{i,j} $ , respectively. Symbols `$\otimes$', `$\odot$' and `$ \times_n $' denote Kronecker product, Khatri-Rao product, mode-$ n $ product, and outer product, respectively, defined as:
\begin{gather*}
	\mathbf{A} \otimes \mathbf{B} \triangleq \begin{bmatrix}
	 a_{11}\mathbf{B}  &  a_{12}\mathbf{B}  &  \cdots \\  
	 a_{21}\mathbf{B}  & a_{22}\mathbf{B}  &  \cdots \\ 
	 \vdots   &  \vdots   &  \ddots  
	\end{bmatrix} , \\
	 \mathbf{A}  \odot  \mathbf{B}  \triangleq  \left [  \mathbf{a}_1  \otimes   \mathbf{b}_1 ,  \mathbf{a}_2  \otimes   \mathbf{b}_2, \cdots \right ] ,\\
	 \left ( \boldsymbol{\mathcal{T}} \times_n  \mathbf{G} \right )_{i_1 ...i_{n - 1} ,j,i_{n + 1} ...i_N } \triangleq \sum _n t_{ i_1 ... i_N} g_{j ,i_n},\\
	 \left ( \mathbf{a} \circ \mathbf{b} \circ \mathbf{c}  \right )_{i,j,k} \triangleq a_ib_jc_k.
\end{gather*}

  For the mode-$ n $ product we assume that the $ n $th dimension of $ \boldsymbol{\mathcal{T}} $ is equal to the number of columns of $ \mathbf{G} $.  Transpose,  Moore-Penrose pseudo inverse, and Frobenius norm are denoted as $ \left ( \cdot \right )^T,\left ( \cdot \right )^\dagger,\left \| \cdot \right \|_F $ respectively. \textsc{Matlab} notations will be used to denote submatrices of a tensor. For instance, we use $ \boldsymbol{\mathcal{T}}\left ( :,:,k\right ) $ to denote the frontal slice of a tensor by fixing the third index to $ k $.  
  
  The mode-$ n $ vectors of  $ \boldsymbol{\mathcal{T}} $ are obtained by fixing all but the $ n $th index of  $ \boldsymbol{\mathcal{T}} $. For a given matrix $ \mathbf{T}
  \in \mathbb{R}^{I \times J} $, $ {\rm vec}\left ( \mathbf{T}\right ) \triangleq\left [ t_1^T,\cdots, t_I^T\right ] \in \mathbb{R}^{IJ} $ denotes column-wise vectorization of $ \mathbf{T}
   $ and $ {\rm unvec}\left ( \cdot \right ) $ performs the inverse. The mode-$ n $ matricization of a third-order tensor $  \boldsymbol{\mathcal{T}} \in \mathbb{R}^{I \times J \times K} $ is denoted as $ \mathbf{T}_n, n = 1,2,3 $, and defined by:
\begin{equation*}
(\mathbf{T}_1)_{i, (j-1)K+k} = (\mathbf{T}_2)_{j, (i-1)K+k} = (\mathbf{T}_3)_{k, (i-1)J+j} = t_{i,j,k}.
\end{equation*}

 A polyadic decomposition (PD) of $  \boldsymbol{\mathcal{T}} $ expresses $  \boldsymbol{\mathcal{T}} $ as the sum of rank-1 terms:
 \begin{equation*}
 \boldsymbol{\mathcal{T}} = \left[\kern-0.15em\left[ \mathbf{A},\mathbf{B},\mathbf{C} \right]\kern-0.15em\right]_R = \sum_{r=1}^{R}\mathbf{a}_r \circ \mathbf{b}_r \circ \mathbf{c}_r\in \mathbb{R}^{I\times J \times K}.
 \end{equation*}
where $ \mathbf{A} \triangleq \left [ \mathbf{a}_1,\cdots ,\mathbf{a}_R \right ] \in \mathbb{R}^{I \times R}  $, $ \mathbf{B} \triangleq \left [ \mathbf{b}_1,\cdots ,\mathbf{b}_R \right ] \in \mathbb{R}^{J \times R} $, and $ \mathbf{C} \triangleq \left [ \mathbf{c}_1,\cdots ,\mathbf{c}_R \right ] \in \mathbb{R}^{K \times R} $. We call it a canonical PD (CPD) if $ R $ is minimal, and the $ R $  is the rank of $ \boldsymbol{\mathcal{T}} $.

\section{Problem Formulation}
We consider the computation of an approximate CPD of a large-scale third-order tensor $  \boldsymbol{\mathcal{T}} $ of size  $ I \times J \times K $. The term ``approximate'' indicates that the CPD model does not hold precisely for $ \boldsymbol{\mathcal{T}} $, due to practical model defect such as model mismatch or noise, and we formulate this model defect as an additive noise term to an exact CPD model. That is to say, the data tensor $ \boldsymbol{\mathcal{T}} $ is formulated as a structured tensor $ \boldsymbol{\mathcal{S}} $  plus an unstructured noise tensor $ \boldsymbol{\mathcal{N}} $:
\begin{equation}\label{CPDNoise}
\boldsymbol{\mathcal{T}} =  \boldsymbol{\mathcal{S}} +  \boldsymbol{\mathcal{N}},
\end{equation}
where $ \boldsymbol{\mathcal{S}} $ admits a third-order CPD of rank $ R $:
\begin{equation}\label{CPDNoise2}
	\boldsymbol{\mathcal{S}} = \left[\kern-0.15em\left[ \mathbf{A},\mathbf{B},\mathbf{C} \right]\kern-0.15em\right]_R.
\end{equation}
with $\mathbf{A}\in \mathbb{R}^{I \times R}, \mathbf{B}\in \mathbb{R}^{J \times R}, \mathbf{C}\in \mathbb{C}^{K \times R}  $ being the first, second, and third factor matrices, respectively.

We assume that the data tensor has large size and low rank, that is to say, ${\rm min}(I, J, K) \gg  R$. We also assume that the factor matrices $ \mathbf{A}, \mathbf{B}, \mathbf{C} $ all have full column rank, and the approximate CPD problem in this case is labelled as  overdetermined. Note that we do not consider the underdetermined case where one or more factor matrices do not have full column rank. We can see from (2) that  $ \boldsymbol{\mathcal{S}} $ has exact multilinear ${\rm rank}\  (R, R, R) $, i.e., $ R_n = {\rm rank}\left ( \mathbf{S}_n\right ) = R, n = 1, 2, 3 $.

As such, the approximate CPD model (\ref{CPDNoise}) can be written in the following Tucker form:
\begin{equation}\label{TuckerForm}
	\boldsymbol{\mathcal{T}} = \boldsymbol{\mathcal{I}}\times_1 \mathbf{A} \times_2 \mathbf{B} \times_3 \mathbf{C} + \boldsymbol{\mathcal{N}}.
\end{equation}

The problem of approximate CPD can be formulated as the following least squares (LS) based optimization problem:
\begin{equation}\label{LSFunction}
	\left \{\tilde{\mathbf{A}}, \tilde{\mathbf{B}}, \tilde{\mathbf{C}}\right \} = \underset{\mathbf{A}, \mathbf{B},\mathbf{C}}{\arg \min} \left ( \boldsymbol{\mathcal{T}} - \boldsymbol{\mathcal{I}}\times_1 \mathbf{A} \times_2 \mathbf{B} \times_3 \mathbf{C}\right ).
\end{equation}
where $ \tilde{\mathbf{A}} $, $ \tilde{\mathbf{B}} $, $ \tilde{\mathbf{C}} $ denote the estimates of $ \mathbf{A} $, $ \mathbf{B} $, $ \mathbf{C} $, respectively.

\section{ Proposed Algorithm }

\subsection{CPD Based on Tensor Compression}

Let us assume that the structured part $ \boldsymbol{\mathcal{S}} $ in the data tensor has exact rank $ R $, and hence multilinear ${\rm rank}\  (R, R, R) $, as explained in Section II. We construct unitary matrices $ \mathbf{U}, \mathbf{V}, \mathbf{W} $. The columns of each matrix form a set of orthonormal bases of the vector space in corresponding mode.

In practice, noise is usually present and we do not directly know $ \boldsymbol{\mathcal{S}} $. In this case, $ \mathbf{U}, \mathbf{V}, \mathbf{W} $ need to be estimated from the noisy tensor $ \boldsymbol{\mathcal{T}} $. Methods for estimating these matrices will be briefly explained later. Note that we often let $ \mathbf{U}, \mathbf{V}, \mathbf{W} $ have more than $ R $ columns to oversample the vector space of each tensor mode, such that the major structure of the tensor in that mode is well preserved. We denote the number of columns of $ \mathbf{U}, \mathbf{V}, \mathbf{W} $ as  $ R' $, $ R' > R $. 

We can use $ \mathbf{U}, \mathbf{V}, \mathbf{W} $ to project the tensor $ \boldsymbol{\mathcal{T}} $  in all modes into a compressed core tensor $ \boldsymbol{\mathcal{G}} \in \mathbb{R}^{R' \times R' \times R'} $ as follows:
\begin{equation}\label{III.A1}
	\boldsymbol{\mathcal{G}} \triangleq \boldsymbol{\mathcal{T}} \times_1 \mathbf{U}^T \times_2 \mathbf{V}^T \times_3 \mathbf{W}^T.
\end{equation}

Substituting (\ref{TuckerForm}) into (\ref{III.A1}) yields:
\begin{equation}\label{III.A2}
	\boldsymbol{\mathcal{G}} = \boldsymbol{\mathcal{I}} \times_1 \mathbf{A'} \times_2 \mathbf{B'} \times_3 \mathbf{C}' + \boldsymbol{\mathcal{N}}'.
\end{equation}
where $ \mathbf{A'} \triangleq \mathbf{U}^T\mathbf{A} \in \mathbb{R}^{R'\times R}, \mathbf{B'} \triangleq \mathbf{V}^T\mathbf{B} \in \mathbb{R}^{R'\times R}, \mathbf{C'} \triangleq \mathbf{W}^T\mathbf{C} \in \mathbb{R}^{R'\times R},  $ and $ \boldsymbol{\mathcal{N}}' = \boldsymbol{\mathcal{N}} \times_1 \mathbf{U}^T \times_2 \mathbf{V}^T \times_3 \mathbf{W}^T$.

We note that the core tensor $ \boldsymbol{\mathcal{G}} $ also admits an approximate CPD but with smaller size than  $ \boldsymbol{\mathcal{T}} $. Therefore, by computing the approximate CPD of $ \boldsymbol{\mathcal{G}} $ we obtain estimates of its factor matrices $ \mathbf{A}', \mathbf{B}' $  and $ \mathbf{C}' $. Then the factor matrices $ \mathbf{A}, \mathbf{B} $ and $ \mathbf{C} $ can be computed as:	
\begin{equation}\label{III.A3}
	\mathbf{A} = \mathbf{U}\mathbf{A'}, \mathbf{B} = \mathbf{V}\mathbf{B'}, \mathbf{C} = \mathbf{W}\mathbf{C'}.
\end{equation}

We note that the above idea of compression based CPD has been adopted in several existing works such as the classical complex parallel factor analysis (COMFAC) algorithm\cite{bro1999fast} and the CPD function in Tensorlab 3.0 software package\cite{tensorlab3.0}. In these works, the projection matrices $ \mathbf{U}, \mathbf{V} $ and $ \mathbf{W} $ are computed via low multilinear rank approximation (LMLRA), which is usually computed iteratively, i.e., via the higher-order orthogonal iteration. However, a truncated multilinear singular value decomposition (MLSVD) may still offer near-to-optimal results and thus is often adopted to compute LMLRA.

\subsection{Random Projection for Tensor Compression}
For a large-scale tensor with low rank, the LMLRA based tensor compression becomes computationally prohibited. Therefore, RAP based methods were proposed. We note that RAP was originally used to compute the SVD of a large-scale matrix\cite{woolfe2008fast, rSVD2019} and was later adapted for the computation of various large-scale tensor models\cite{sidiropoulos2014parallelA,erichson2020randomized,yuan2019randomized,che2019randomized,minster2020randomized}. Here we explain RAP with power iterations  proposed in \cite{erichson2020randomized}.

First, a sketch matrix is constructed  that holds a set of approximate base vectors of the vector space of each mode of the tensor as follows (we take the construction of the sketch matrix in the first mode as an example):
\begin{equation}\label{III.B1}
\mathbf{X}^{(m)} = \mathbf{T}_1^{(m)}\mathbf{\Psi }_1.
\end{equation}
where $ \mathbf{T}_1^{(m)} $ is defined as: $ \mathbf{T}^{\left ( m\right )}_1 \triangleq  \left ( \mathbf{T}_1\mathbf{T}_1^T\right )^m\mathbf{T}_1 $, $ \mathbf{T}_1 $ is the mode-1 matricization of tensor $ \boldsymbol{\mathcal{T}} $ and $ \mathbf{\Psi }_1 $ is a random test matrix used to sample the mode-1 vector space of $ \boldsymbol{\mathcal{T}} $, i.e., the column space of $ \mathbf{T}_1 $. Note that the randomly generated columns of $ \mathbf{\Psi }_1 $ are linearly independent in the generic sense, and that  $ \mathbf{T}_1^{(m)} $ and $ \mathbf{T}_1 $ have identical column space. Hence, the set of column vectors $ \left \{\mathbf{x}_1,\cdots ,\mathbf{x}_{R'} \right \} $ of $ \mathbf{X}^{(m)} $ provides an efficient sample of the mode-1 vector space of $ \boldsymbol{\mathcal{T}} $. In the noiseless case, an economic QR decomposition of the sketch matrix $ \mathbf{X}^{(m)} $ provides a set of orthonormal bases of that vector space and thus the projection matrix $ \mathbf{U}^{(m)} $ is constructed using these basis vectors as columns. When noise is present, we can perform a truncated SVD(T-SVD) of the sketch matrix to estimate $ \mathbf{U}^{(m)} $. Note that we use $ \mathbf{T}_1^{(m)} $ instead of $ \mathbf{T}_1 $ to construct $ \mathbf{X}^{(m)} $ because the singular values of $ \mathbf{T}_1^{(m)} $ are $ \left \{\sigma_r^{2m+1}\right \} $ with $ \left \{\sigma_r\right \} $ being the singular values of $ \mathbf{T}_1 $ such that the singular vectors of $ \mathbf{T}_1^{(m)} $ associated with major singular values are more identifiable.

 In practice, the sketch matrices are usually constructed via normalized subspace iterations\cite{erichson2020randomized}. After the construction of the sketch matrices, $ \mathbf{X}^{(m)} $ and $ \mathbf{Y}^{(m)} $, $ \mathbf{Z}^{(m)} $ that correspond to the first, second, and third mode of the tensor, respectively, the projection matrices $ \mathbf{U}^{(m)}, \mathbf{V}^{(m)}, \mathbf{W}^{(m)} $ are  computed via the T-SVD of $ \mathbf{X}^{(m)} $ and $ \mathbf{Y}^{(m)} $, $ \mathbf{Z}^{(m)} $ , respectively.

\subsection{Coupled Random Projection Based CPD}
Now we propose a CPD algorithm for large-scale tensors based on coupled RAP (CoRAP). First, we construct sketch matrices $ \mathbf{X}^{(m)} $ and $ \mathbf{Y}^{(m)} $ in the first and second mode, respectively, according to (\ref{III.B1}), where $ m $ is a fixed integer denoting the order of power iterations. Performing the T-SVD of $ \mathbf{X}^{(m)} $ and  $ \mathbf{Y}^{(m)} $ yields the projection matrices  $ \mathbf{U}^{(m)} \in \mathbb{R}^{I \times R'} $ and  $ \mathbf{V}^{(m)} \in \mathbb{R}^{I \times R'} $, respectively.

Then we let $ m $ vary and repeat the above procedure for each $ m $. As such, we obtain two sets of projection matrices $ \left \{ \mathbf{U}^{(m)}\right \},\left \{ \mathbf{V}^{(m)}\right \} $ in the first and second mode, respectively, where $ m = 1,...,M $ and $ M $ is the maximal order of power iterations. In the third mode, we construct one sketch matrix $ \mathbf{Z} $ and  calculate the projection matrix  $ \mathbf{W} \in \mathbb{R}^{K \times R'} $.

Finally, we obtain $ M $ projection matrices in the first mode and the second mode, respectively, as well as one projection matrix in the third mode. We group all the projection matrices into $ M $ triads: $ \boldsymbol{\Omega} ^{(m)} \triangleq \left \{ \mathbf{U}^{(m)}, \mathbf{V}^{(m)}, \mathbf{W} \right \},\  m = 1, ..., M $. Each triad $ \boldsymbol{\Omega} ^{(m)} $ contains projection matrices in all three modes, while different triads share a common projection matrix $ \mathbf{W} $ in the third mode.

We perform the following projection using each $ \boldsymbol{\Omega} ^{(m)} $ (\ref{III.A1}):
\begin{equation}\label{III.C1}
\boldsymbol{\mathcal{G}}^{(m)} \triangleq \boldsymbol{\mathcal{T}} \times_1 \mathbf{U}^{(m)T} \times_2 \mathbf{V}^{(m)T} \times_3 \mathbf{W}^T,
\end{equation}

According to (\ref{III.A2}), each $ \boldsymbol{\mathcal{G}}^{(m)} $ admits an approximate CPD:
\begin{equation}\label{III.C2}
\boldsymbol{\mathcal{G}}^{(m)} = \left[\kern-0.25em\left[ \mathbf{A}^{(m)},\mathbf{B}^{(m)},\mathbf{C}' \right]\kern-0.25em\right] + \boldsymbol{\mathcal{N}}^{(m)}.
\end{equation}
where  $\, \mathbf{A}^{(m)} \triangleq \mathbf{U}^{(m)T}\mathbf{A}, \mathbf{B}^{(m)} \triangleq \mathbf{R}^{(m)T}\mathbf{B}, \mathbf{C}' \triangleq \mathbf{W}^T\mathbf{A}  $ and $\, \boldsymbol{\mathcal{N}}^{(m)} \triangleq \boldsymbol{\mathcal{N}} \times_1 \mathbf{U}^{(m)T} \times_2 \mathbf{V}^{(m)T} \times_3 \mathbf{W}^{T}  $.

Note that the set of tensors $ \{\boldsymbol{\mathcal{G}}^{(m)}, m=1,...,M\} $  together admits an approximate C-CPD, with a common factor matrix  $\mathbf{C}' $ in the third mode. Therefore, instead of performing the uncoupled CPD for each tensor  $ \boldsymbol{\mathcal{G}}^{(m)} $, we can perform a C-CPD for all the tensors  $ \{\boldsymbol{\mathcal{G}}^{(m)} \}  $. The projections (\ref{III.C1}) for all $ m = 1,...,M $  are together labelled as CoRAP.

As long as  $ \mathbf{A}^{(m)} $,  $ \mathbf{B}^{(m)} $, and $ \mathbf{C}' $  are computed by  C-CPD of  $  \{\boldsymbol{\mathcal{G}}^{(m)} \} $,  $ m = 1,...,M $, we recover the factor matrices $ \mathbf{A}, \mathbf{B} $ and $ \mathbf{C} $ by (\ref{III.A3}). Note that as $ m $ varies from 1 to $ M $, we indeed have $ M $ estimates of $ \mathbf{A} $ and $ \mathbf{B} $, respectively. Here we select the optimal value of $ m $, $ m_{opt} $, such that  $ \left[\kern-0.25em\left[ \mathbf{A}^{(m_{opt})},\mathbf{B}^{(m_{opt})},\mathbf{C}' \right]\kern-0.25em\right] $ offers the best fit of the data tensor  in the least squares sense:
\begin{equation}\label{III.C3}
m_{opt} = \underset{m}{\arg \min} \left \| \boldsymbol{\mathcal{G}}^{(m)} - \left[\kern-0.25em\left[ \mathbf{A}^{(m)},\mathbf{B}^{(m)},\mathbf{C}' \right]\kern-0.25em\right] \right \|^2_F.
\end{equation}

The factor matrices of the original tensor $ \boldsymbol{\mathcal{T}} $ are then computed as: 
\begin{equation}\label{III.C4}
\, \mathbf{A} \!=\! \mathbf{U}^{(m_{opt})}\mathbf{A}^{(m_{opt})}, \mathbf{B} \!=\! \mathbf{V}^{(m_{opt})}\mathbf{B}^{(m_{opt})}, \mathbf{C} \!=\! \mathbf{W}\mathbf{C}'.
\end{equation}

\subsection{Coupled CPD}
As explained in Subsection III.C, C-CPD plays an important role in CoRAP based CPD. In fact, C-CPD algorithms have been studied tremendously in the past decade, including the algebraic algorithms\cite{sorensen2015coupledB,gong2019double,gong2018double} and optimization based algorithms\cite{sorber2015structured,gong2018double}, which are shown to have their own pros and cons, respectively. In particular, the algebraic algorithms are guaranteed to return the exact solution in the noiseless case. However, they only return suboptimal solution when noise is present. On the other hand, the optimization based algorithms can provide optimal results in the LS sense if they converge to the global minimum. However, they are likely to converge to a local minimum and are thus sensitive to initialization. As such, we can use the algebraic algorithms to efficiently initialize the optimization based algorithms.

We note that the C-CPD problem in our setting (\ref{III.C2}) is overdetermined in the sense that all the factor matrices have full column rank. Therefore, instead of using the algebraic algorithms in \cite{sorensen2015coupledB} and \cite{gong2019double} that are designed for underdetermined problems, we can use a simple algebraic C-CPD algorithm, summarized as follows:
\begin{itemize}
	\item [1)] 
		Calculate the CPD of one tensor $ \boldsymbol{\mathcal{G}}^{(m)} $ with fixed $ m $. We assume without loss of generality that CPD of $ \boldsymbol{\mathcal{G}}^{(1)} $ is performed to compute estimates of  $ \mathbf{A}^{(1)} $,  $ \mathbf{B}^{(1)} $, and  $ \mathbf{C}' $;      
	\item [2)]
		Since $ \mathbf{C}' $ is already obtained, we have: 
	\begin{equation}\label{III.D1}
	\mathbf{G}_3^{(m)} \left ( \mathbf{C}'^T\right )^\dagger = \mathbf{A}^{(m)}\odot \mathbf{B}^{(m)} + \mathbf{N}'^{(m)},
	\end{equation}
	where  $ m \in \left [ 2, M\right ] $, and  $ \mathbf{N}'^{(m)} $ is the noise term. It is shown in (16) that each column of  $ \mathbf{G}_3^{(m)} \left ( \mathbf{C}'^T\right )^\dagger $, denoted as  $ \mathbf{g}_r^{(m)} $, is approximately a vectorized rank-1 matrix:
	\begin{equation}\label{III.D2}
	{\rm unvec}(\mathbf{g}_r^{(m)}) \approx \mathbf{a}_r^{(m)}\cdot \mathbf{b}_r^{(m)T}, \ r = 1,...,R.
	\end{equation}
	Therefore, by performing rank-1 approximation of $ {\rm unvec}(\mathbf{g}_r^{(m)}) $  with varying $ r $ and fixed $ m $ we obtain factor matrices $ \mathbf{A}^{(m)} $ and  $ \mathbf{B}^{(m)} $. We let $ m $ vary and repeat the above procedure (\ref{III.D1}) and (\ref{III.D2}) for each $ m $, then all the factor matrices can be computed.
\end{itemize}

Now that we obtain the initial results via an algebraic algorithm, we use them as initialization of an optimization based C-CPD algorithm to obtain the final results. Possible candidates include the alternating least squares (ALS)\cite{gong2018double} and nonlinear least squares (NLS)\cite{sorber2013optimization} algorithms.

\section{Numerical Experiments}

In this section we use numerical experiments to illustrate the performance of the proposed CoRAP based CPD algorithm in comparison with RAP based CPD \cite{erichson2020randomized} and direct CPD. In all the experiments the data tensor is generated as follows:
\begin{equation}\label{TenCons}
\boldsymbol{\mathcal{T}} = P_s \boldsymbol{\mathcal{S}} + P_n \boldsymbol{\mathcal{N}}_n,
\end{equation}
where tensor $ \boldsymbol{\mathcal{S}} $ is constructed according to (\ref{CPDNoise2}) with the factor matrices $ \mathbf{A} $, $ \mathbf{B} $, $ \mathbf{C} $ randomly drawn from a Gaussion distribution with zero mean and unit variance. $ \boldsymbol{\mathcal{N}} $ denotes the additive noise term that is also drawn from a Gaussion distribution with zero mean and unit variance. Parameters $ P_s $ and $ P_n $ denote signal level and noise level, respectively. The Signal-to-noise ratio(SNR) is defined as:
\begin{equation}\label{SNRDefin}
\rm SNR \triangleq 10 \rm log_{10}(\textit{P}_\textit{s}/\textit{P}_\textit{n}),
\end{equation}

We evaluate the performance of compared algorithms in terms of mean relative error, defined as follows:
\begin{equation}\label{MRE}
\epsilon  \triangleq \frac{1}{3} \sum _{\mathbf{H}\in \{\mathbf{A},\mathbf{B},\mathbf{C}\}}  \left \|\mathbf{H} \right \|_F^{-2} \left \| \mathbf{H} - \tilde{\mathbf{H}} \mathbf{P} \mathbf{S} \right \|_F^2.
\end{equation}
where $ \mathbf{H} $ and $ \tilde{\mathbf{H}} $ denote the true factor matrix and its estimate, respectively. $ \mathbf{P} $ and $ \mathbf{S} $ are permutation and scaling matrices, respectively, such that the estimated factor matrix $\tilde{\mathbf{H}}$ is permuted and scaled to optimally fit $\mathbf{H}$ in the LS sense. In addition, we use CPU time to evaluate how fast the algorithms completes the calculation. The mean relative error and CPU time are calculated as the average of the results of 200 Monte Carlo runs. Note that the direct CPD is implemented using `cpd.m' function in Tensorlab 3.0 with default setting\cite{tensorlab3.0}.

\begin{figure}[htb]
	\centering
	\includegraphics[width=4.37cm]{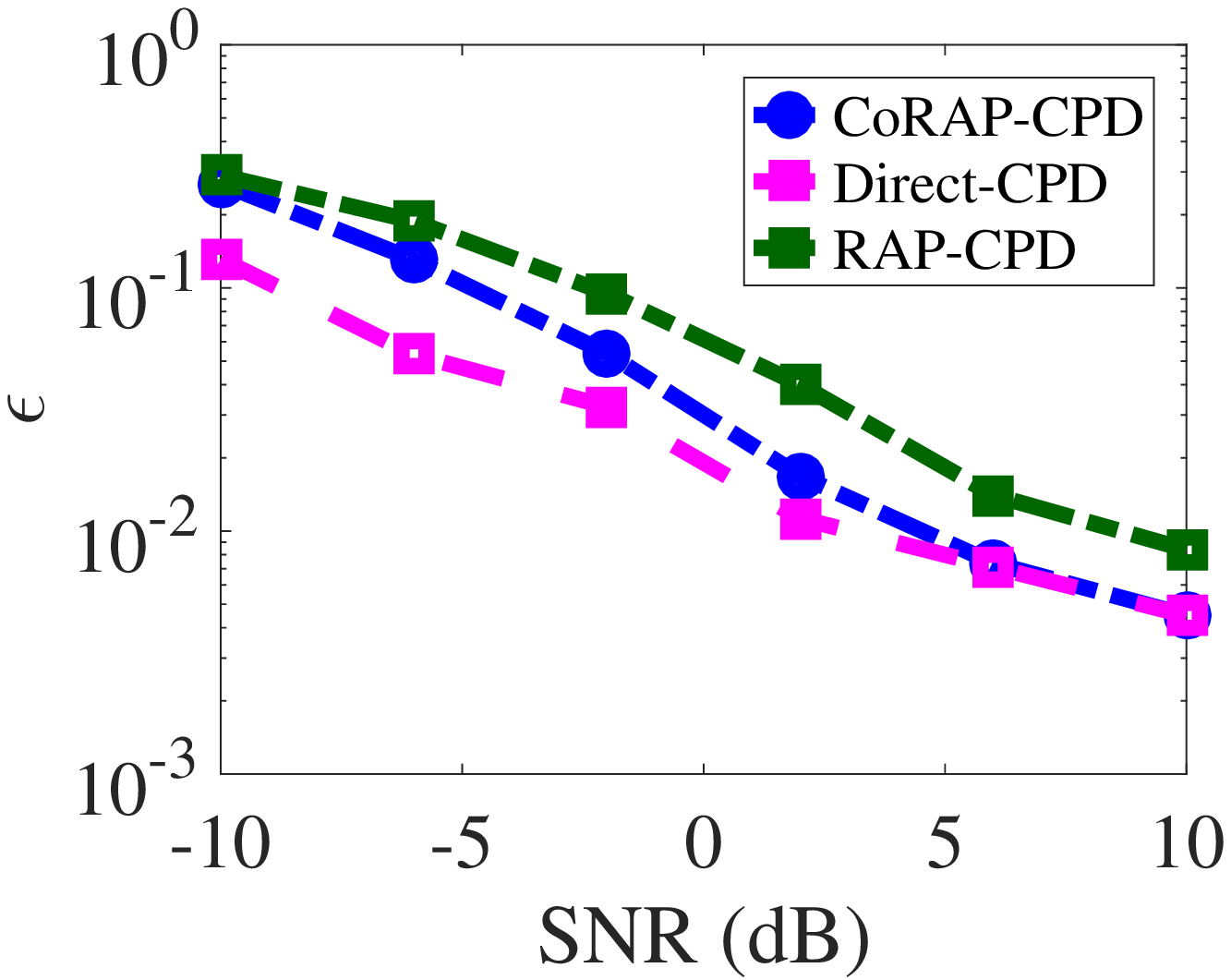}
	\includegraphics[width=4.37cm]{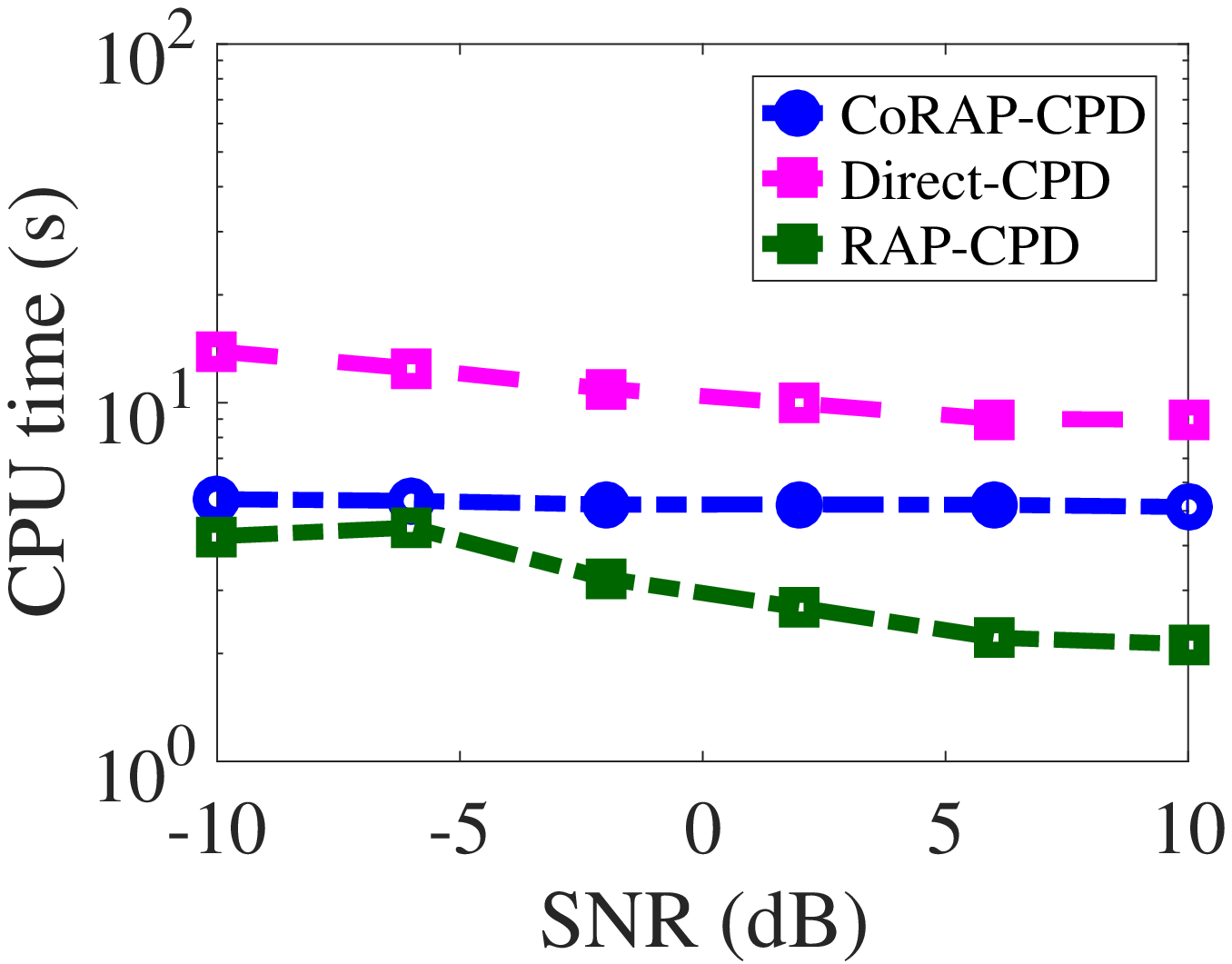}
	\caption{The performance of CoRAP-CPD, RAP-CPD and direct CPD versus SNR of a noisy tensor $ \boldsymbol{\mathcal{T}} $ of size $ 500 \times 500 \times 500 $ and rank $ R = 50 $. The left
		sub-figure shows the overall  error, and the right shows the CPU time}
	\label{fig:pictures2}
\end{figure}

In the first experiment, we mainly study the impact of noise on the compared algorithms. The data tensor $ \boldsymbol{\mathcal{S}} $ is of size $ 500 \times 500 \times 500 $ with rank $ R = 50 $. 
For the proposed algorithm, we set $ M = 2$. In addition, for the implementation of C-CPD in CoRAP-CPD, we use algebraic C-CPD to efficiently initialize an optimization based C-CPD algorithm.  Experiment results, as demonstrated in Fig. 1, show that the performance of CoRAP-CPD, in terms of mean relative error, is better than that of RAP-CPD, and  it approaches that of direct CPD as SNR increases.  On the other hand, the CPU time of CoRAP-CPD is slightly higher than that of RAP-CPD, and is still lower than that of direct CPD. This is because more data tensors are generated and processed for CoRAP than RAP. The observation in this experiment generally suggests that CoRAP based CPD improves the accuracy of its RAP based counterpart, while can still maintain the computational efficiency of RAP, in comparison with direct CPD.

\begin{figure}[htb]
	\centering
	\includegraphics[width=4.37cm]{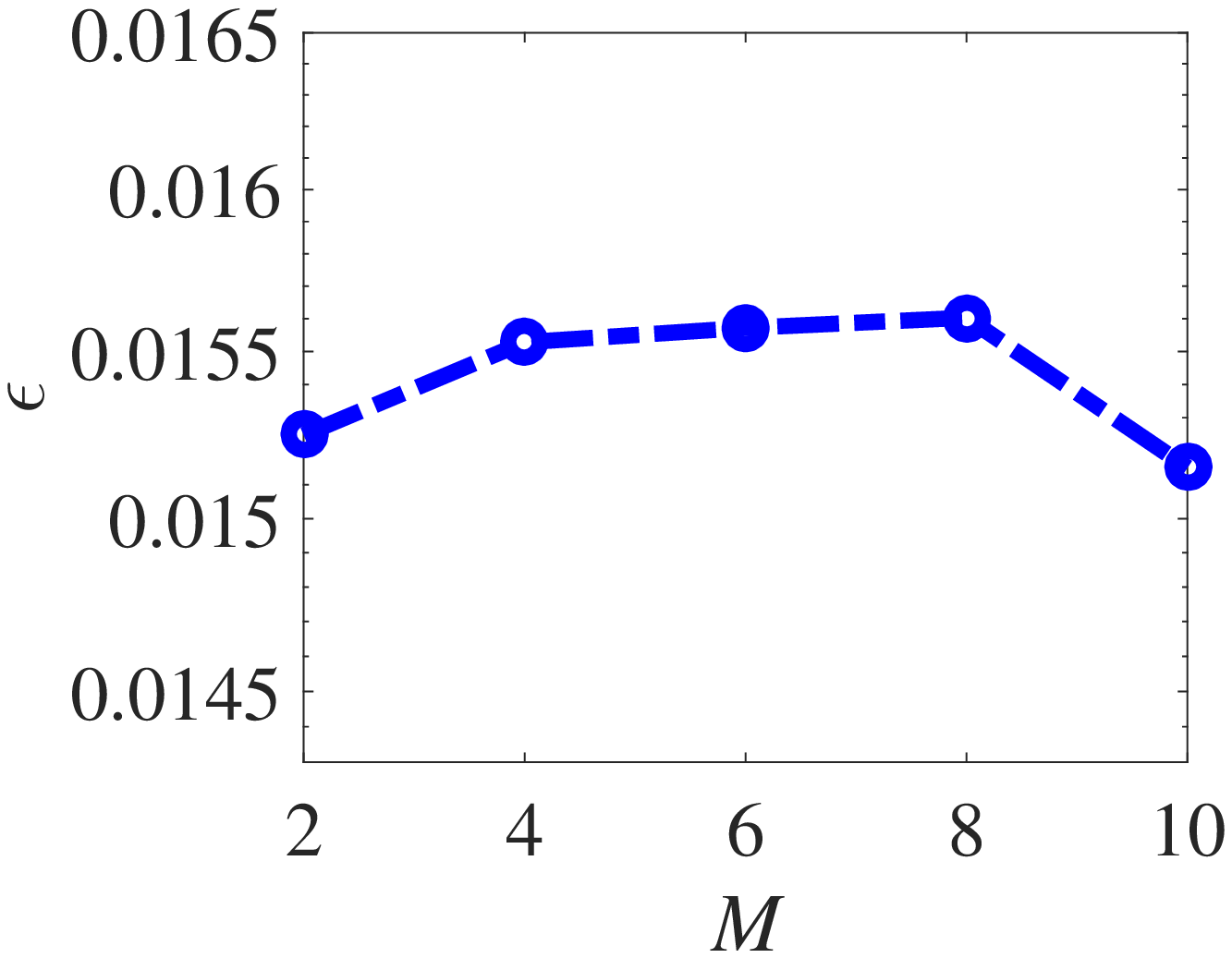}
	\includegraphics[width=4.37cm]{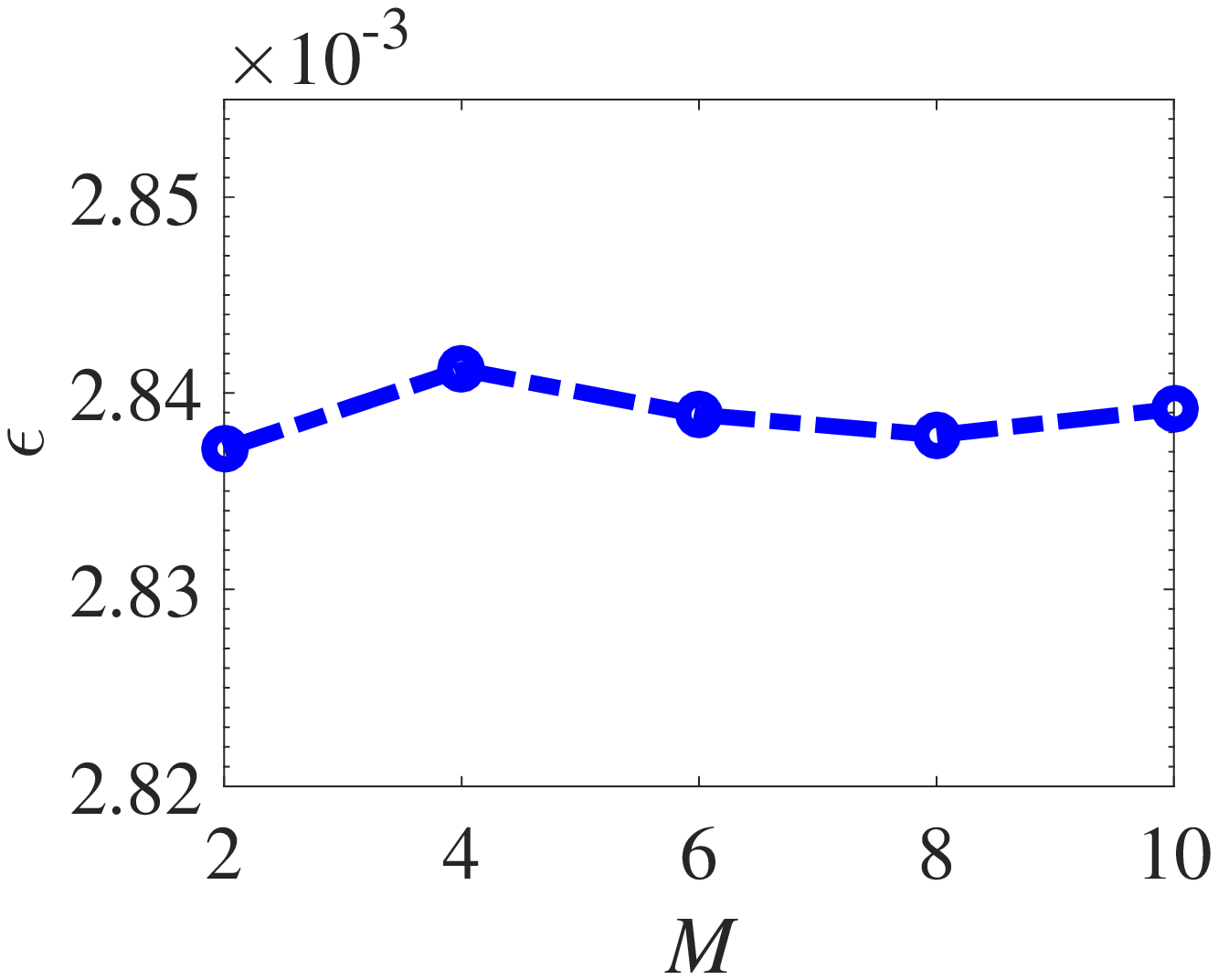}
	\caption{Performance of CoRAP-CPD for different $ M $ of a noisy tensor of size $ 200 \times 200 \times 200 $ with rank $ R = 20 $. The left and right sub-figures correspond to SNR = $ -2 $dB and SNR = 10dB, respectively.}
	\label{fig:pictures2}
\end{figure}

In the second experiment, we study the impact of $ M $ on the accuracy of CoRAP-CPD. The data tensor is generated analogously to the last experiment except that we let $ M $ vary while SNR is fixed. The results for SNR = $ -2 $dB and SNR = 10dB are given in Fig. 2. The observation generally suggests that  $ M $ has little impact on the accuracy of CoRAP-CPD, while a larger $ M $ would lead to higher complexity. Therefore, in practice we can generally use a small $ M $, i.e.,$  M = 2 $.

\begin{figure}[htb]
	\centering
	\includegraphics[width=4.37cm]{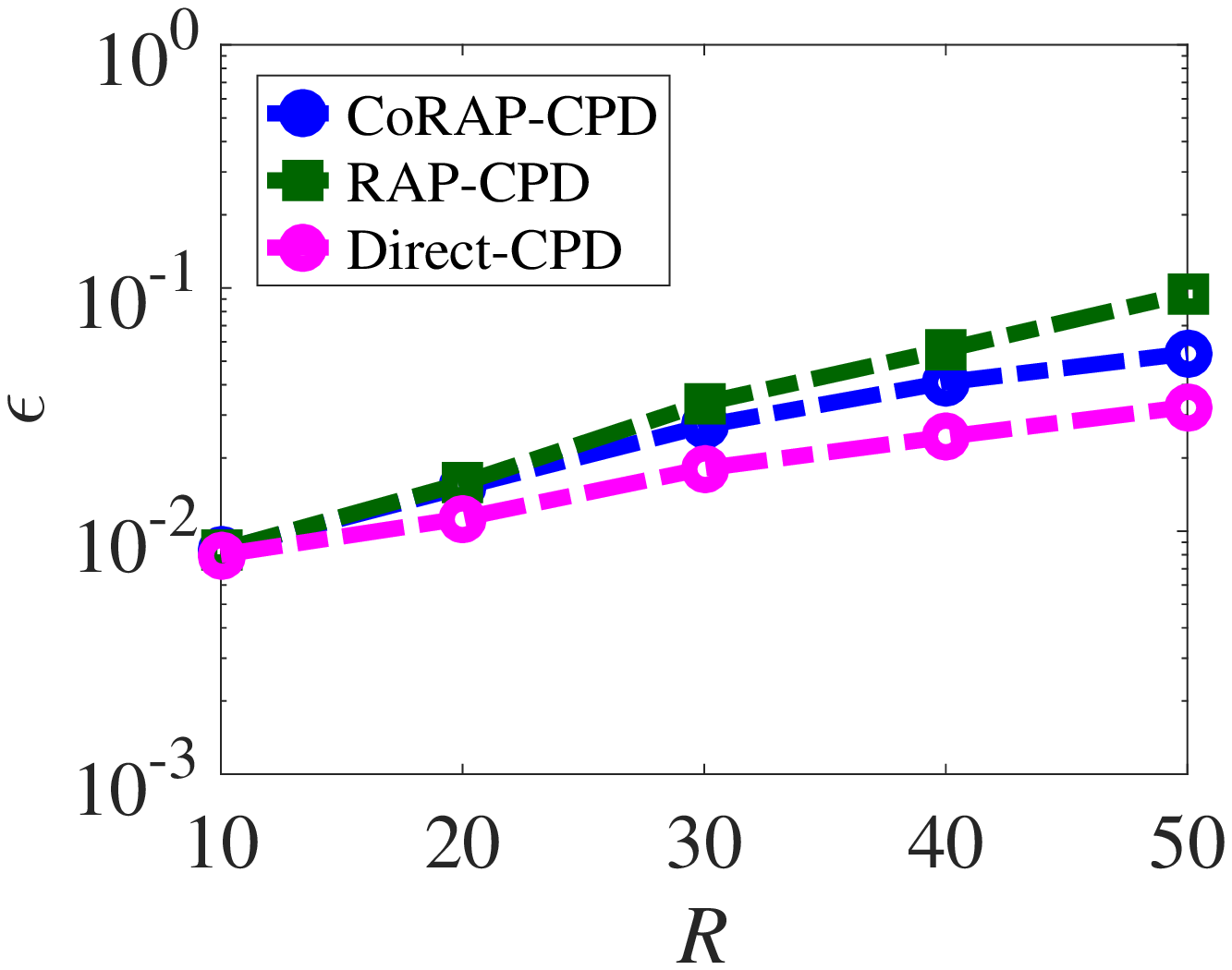}
	\includegraphics[width=4.37cm]{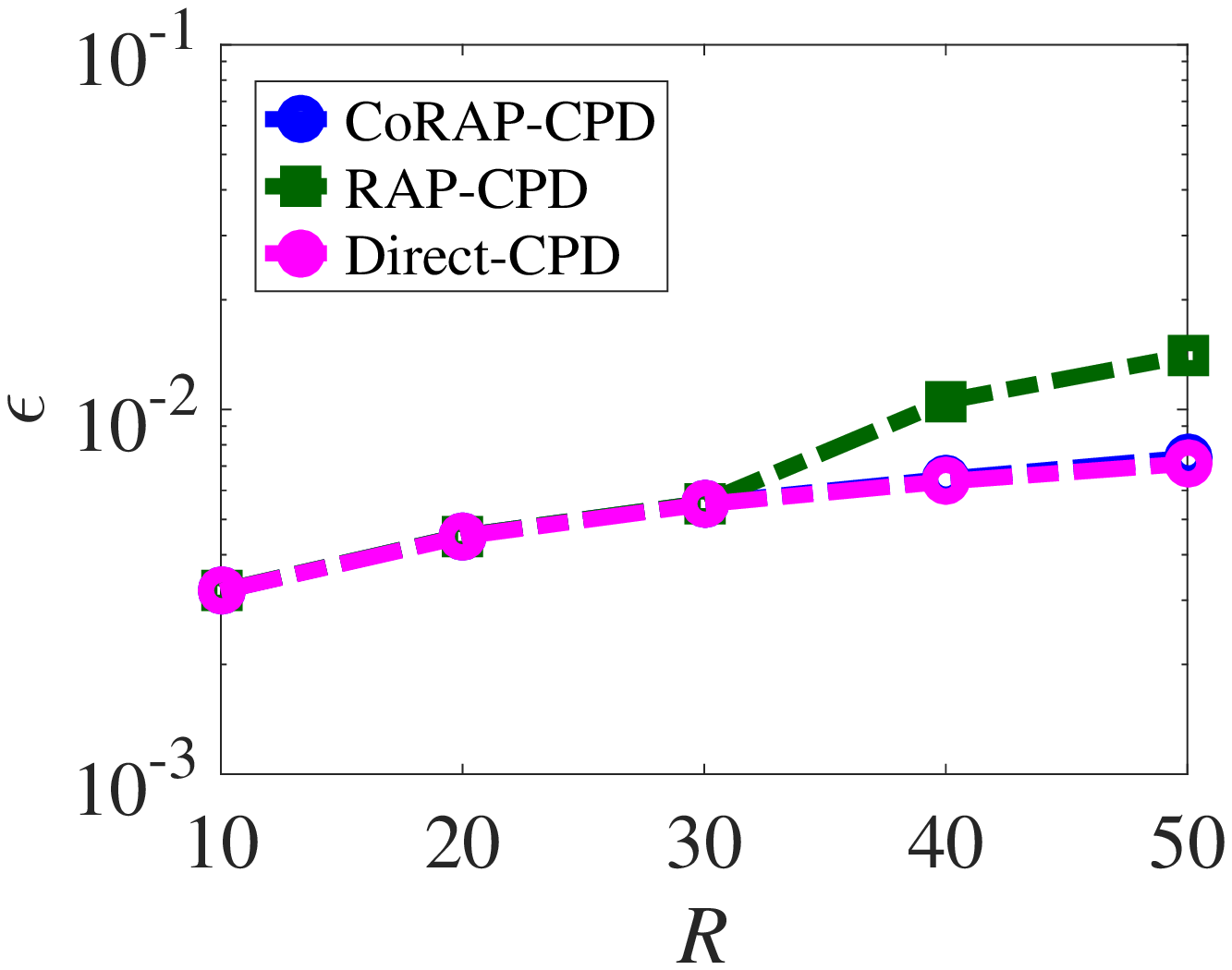}
	\caption{The performance of CoRAP-CPD, RAP-CPD and direct CPD versus rank $ R $ of a noisy tensor $ \boldsymbol{\mathcal{T}} $ of size $ 500 \times 500 \times 500 $. The left and right sub-figures correspond to SNR = $ -2 $dB and SNR = 6dB, respectively.}
	\label{fig:pictures2}
\end{figure}

In the third experiment, we study the impact of tensor rank, $ R $, on the performance of CoRAP-CPD and RAP-CPD. The data tensor is generated in an analogous manner as the last two experiments. We fix SNR to $ -2 $dB and 6dB, respectively, and let $ R $ vary from 10 to 50. The results are given in Fig. 3. It has been observed that in both low SNR (SNR = $ -2 $dB) and medium SNR (SNR = 6dB), the proposed CoRAP based CPD algorithm has remarkably better accuracy than RAP-CPD when $ R $ is large. This observation clearly shows the interests of the proposed algorithm in difficult cases where the data tensor not only has large size but also relatively high rank.

\section{Conclusion}
A novel CPD algorithm for large-scale tensor is proposed by combining the idea of random projection (RAP) and coupled tensor decomposition. The coupled random projection (CoRAP) technique is developed and applied to project the original large-scale tensor into a set of small tensors that together admits a coupled CPD(C-CPD), and a C-CPD algorithm is used to jointly decompose these tensors. Then, the results of C-CPD are back projected to obtain the factor matrices of the original tensor. Through numerical experiments, we have shown that the proposed CoRAP based CPD algorithm has better performance in terms of accuracy than RAP based CPD, with slightly increased CPU time. In particular, the proposed algorithm has interests in difficult cases where the data tensor not only has large size but also relatively high rank.

\bibliographystyle{IEEEbib}
\bibliography{strings1,refs1,elec}

\end{document}